\documentclass[a4paper,fleqn]{cas-dc}
\usepackage{xcolor}
\hypersetup{
	colorlinks=true,
	linkcolor=Cerulean,
	filecolor=Cerulean,      
	urlcolor=Cerulean,
	citecolor=Cerulean,
}
\usepackage[numbers,sort&compress]{natbib}
\usepackage{subfigure}
\usepackage{multirow}
\usepackage{caption}

\begin{document}
\let\WriteBookmarks\relax
\def\floatpagepagefraction{1}
\def\textpagefraction{.001}

\shorttitle{Y.Ao and H.Wu / Neurocomputing(2021)}

\shortauthors{Y.Ao and H.Wu}

\title [mode = title]{Feature Aggregation and Refinement Network for 2D Anatomical Landmark Detection}


\author[a]{Yueyuan Ao}
\author[a]{Hong Wu\corref{cor1}}
\cortext[cor1]{Corresponding Author}

\address[a]{University of Electronic Science and Technology of China, Chengdu Sichuan, 611731, China}

\nonumnote{\textsl{E-mail address:} \href{mailto:hwu@uestc.edu.cn}{hwu@uestc.edu.cn} (H. Wu)}
\begin{abstract}
Localization of anatomical landmarks is essential for clinical diagnosis, treatment planning, and research. In this paper, we propose a novel deep network, named feature aggregation and refinement network (FARNet), for the automatic detection of anatomical landmarks. To alleviate the problem of limited training data in the medical domain, our network adopts a deep network pre-trained on natural images as the backbone network and several popular networks have been compared. Our FARNet also includes a multi-scale feature aggregation module for multi-scale feature fusion and a feature refinement module for high-resolution heatmap regression. Coarse-to-fine supervisions are applied to the two modules to facilitate the end-to-end training. We further propose a novel loss function named Exponential Weighted Center loss for accurate heatmap regression, which focuses on the losses from the pixels near landmarks and suppresses the ones from far away. Our network has been evaluated on three publicly available anatomical landmark detection datasets, including cephalometric radiographs, hand radiographs, and spine radiographs, and achieves state-of-art performances on all three datasets. Code is available at: \url{https://github.com/JuvenileInWind/FARNet}


\end{abstract}



\begin{keywords}
Anatomical landmark detection \sep Deep network \sep Feature aggregation \sep Feature refinement \sep Exponential Weighted Center loss
\end{keywords}

\maketitle

\section{Introduction} 
\label{sec:introduction}
{A}{natomical} landmark localization is a prerequisite not only for patient diagnosis and treatment planning~\cite{ionasec2008dynamic,zheng2010automatic,wang2015evaluation,wang2016benchmark}, but also for numerous medical image analysis tasks including image registration~\cite{murphy2011semi,2014Robust} and image segmentation~\cite{oktay2016stratified}. In practice, landmarks are usually located manually or semi-manually, which is tedious, time-consuming, and prone to errors. Therefore, there is a strong need for fully automatic and accurate landmark localization approaches. But identifying anatomical landmarks is challenging because of the variations on individual structures, appearance ambiguity, and image complexity. According to the medical image modality used, there exists 2D and 3D anatomical landmark localization. The examples of 2D modality include cephalometric X-Ray images and hand radiographs, and the examples of 3D modality include 3D brain MR scans, 3D olfactory MR scans, and 3D prostate CT images. In this paper, we focus on 2D anatomical landmark localization. 

\begin{figure*}[htbp]
\centering
\includegraphics[scale=0.78]{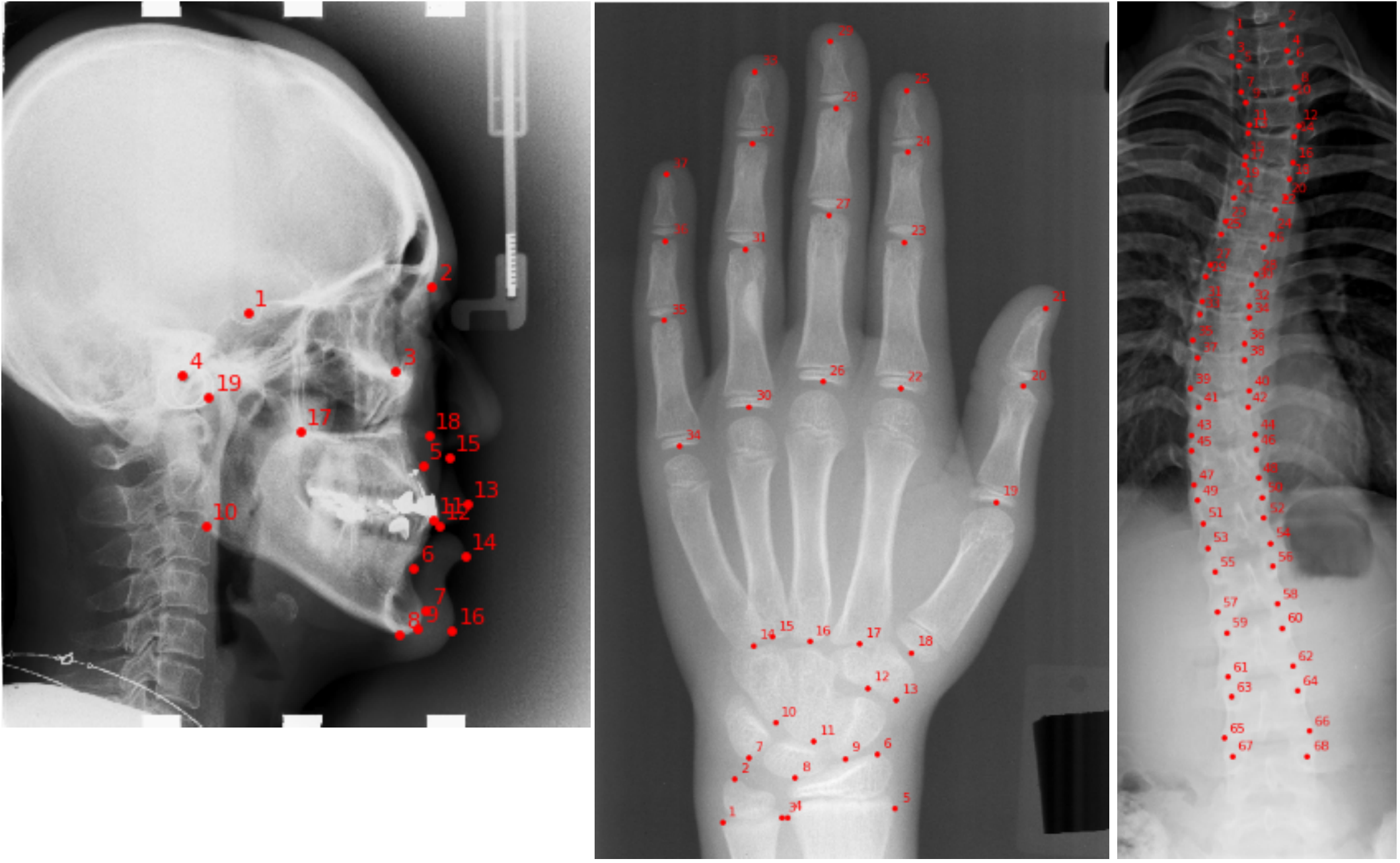}
\caption{Sample images and the anatomical landmarks for three datasets used in this paper. From left to right are lateral cephalogram with 19 landmarks, radiograph of left hand with 37 landmarks and spinal anterior-posterior X-Ray with 68 landmarks.}
\label{example}
\end{figure*}

\setlength{\parskip}{0.1\baselineskip}

\par In the last decades, numerous automatic anatomical landmark detection methods have been proposed. Rule-based methods~\cite{rudolph1998automatic,grau2001automatic} utilize image processing techniques to detect edges/contours and identify the landmarks based on the prior-knowledge of the landmark structures. However, rules would become too complex to be formulated with the increase of the image complexity. Some works adopt template matching~\cite{cardillo1994image, grau2001automatic, el2004automatic} to locate landmarks. To consider both the local appearance and the global spatial configuration of landmarks, some works~\cite{saad2006automatic, yue2006automated, kafieh2007automatic, keustermans2010automated} employ Active Shape Model and Active Appearance Model. Later, machine learning algorithms, such as neural networks, SVM, and random forest, etc., have been applied to landmark localization for better generalization in case of anatomical variation and noise. These methods formulate landmark localization as a classification problem or a regression problem. Classification-based methods~\cite{2003Robust,2005Automatic,criminisi2009decision,zhan2011robust,ibragimov2015computerized} determine whether a landmark is located in an image patch. Regression-based methods~\cite{criminisi2010regression,Criminisi2013Regression,2014Robust,ebner2014towards,lindner2014robust,lindner2015fully,vstern2016local,oktay2016stratified,sun2017direct,urschler2018integrating} predict the displacement from an image patch to a certain landmark. Some machine learning based approaches~\cite{criminisi2009decision,lindner2014robust,ibragimov2015computerized,lindner2015fully,vstern2016local,oktay2016stratified,urschler2018integrating} further combine the local predictions with global configuration modeling to improve the detection accuracy.

\par In recent years, deep learning has achieved great success in many fields such as computer vision and natural language processing and has been widely used in medical image analysis including anatomical landmark detection. Due to the limited medical training data, it is challenging to train deep networks for anatomical landmark detection. To alleviate this problem, some deep learning-based landmark detection methods use local image patches as samples to perform patch-wise regression/classification~\cite{2015Automatic,2016Automatic,arik2017fully}. But it is time-consuming for these methods to train and test on a large number of image patches. Moreover, patch-based methods only utilize local information and ignore global information, thus they are not able to predict all landmarks accurately. Recently, a few end-to-end CNN-based methods have been proposed for landmark detection, which utilize an entire image as input and facilitate the modeling of global information. Some of them~\cite{lee2017cephalometric,2017Automatic} directly regress the landmark coordinates, and most of the others~\cite{riegler2015anatomical,payer2016regressing,lee2017cephalometric,o2018attaining,2019Integrating,zhong2019attention} adopt fully convolutional networks (FCN) to regress heatmaps, each of which encodes the pseudo-probability of a landmark being located at a certain pixel position. However, due to the limited training data, very shallow networks are adopted in these methods and limit their capacities. In addition, the output resolutions of the previous networks usually have stride of 4 pixels with respect to the input image or are even smaller, which further introduce quantization errors to the predictions. Therefore, there's a need for developing deep networks with high-resolution feature extraction for accurate anatomic landmark detection.

In this paper, we propose a novel end-to-end deep network (shown in {\color{Cerulean}Fig.} \ref{farnet}) for anatomic landmark detection. To alleviate the problem of limited training data, we adopts a deep network pre-trained on natural images as the backbone network. A multi-scale feature aggregation (MSFA) module is proposed to combine multi-scale features extracted by the backbone with up-sampling path, down-sampling path, and skip connections. Features with different resolutions are combined by concatenation in a higher-resolution-dominate manner. The elaborate design of the MSFA module achieves a good trade-off between the network capacity and efficiency. To achieve high-resolution heatmap regression, we further propose a feature refinement (FR) module that combines the feature maps convoluted from the input image with the up-sampled feature maps and heat maps from the MSFA module to generate features having the same resolution as the input image. Coarse-to-fine supervisions are also applied to the two modules to facilitate the end-to-end training. To achieve accurate heatmap regression, we further propose a novel loss function named Exponential Weighted Center loss, which focuses on the losses from the pixels near landmarks and suppresses the losses from far away. Our end-to-end network, named Feature Aggregation and Refinement Network (FARNet), is evaluated on three publicly available datasets in the medical domain (examples are shown in {\color{Cerulean}Fig.} \ref{example}), a cephalometric radiograph dataset~\cite{wang2016benchmark}, a hand radiograph dataset~\cite{2019Integrating} and a spinal Anterior-Posterior X-ray dataset~\cite{2017Automatic}. Our network achieves state-of-the-art performances on all these datasets, which proves the effectiveness and generality of our network. 

Our contributions are summarized as follows,
\begin{enumerate}
\item We propose a novel deep network for anatomic landmark detection, which can fuse multi-scale features from the backbone network and achieve high-resolution heatmap regression.
\item We compare several widely-used pre-trained networks as the backbone in our network and DenseNet-121 achieves the best performance.
\item To achieve more accurate localization, we propose a novel loss function named Exponential Weighted Center loss for heatmap regression.
\item Experimental results indicate that our network achieves state-of-art performances on three public medical datasets.
\end{enumerate}

The rest of this paper is organized as follows.  Section II presents a brief review of some important related works. Section III describes our proposed network and loss function in detail. Section IV reports the experimental setups and results on three public medical datasets. Finally, the conclusions are drawn in Section V.

\section{Related Work}

\subsection{Deep Learning Methods for Anatomical Landmark Detection}

\par Deep learning has achieved great success in many computer vision applications and has also been applied to anatomic landmark detection. One of the major challenges to deep learning-based anatomical landmark detection is the limited medical imaging data for network training. To alleviate this problem, some deep learning-based landmark detection methods perform patch-wise regression/classification. But the patch-based approach is time-consuming and not able to capture global information which is also important for accurate prediction. For 2D landmark detection, another solution to the limited training data problem is to use backbone networks pre-trained on natural images. To incorporate long-range/global spatial context with local information for landmark prediction, some methods combine patch-based CNN predictions with a statistical shape model, some adopt CNNs with encoder-decoder structure, others learn global context and local features in different branches or in sequence.

Aubert et al. \cite{2016Automatic} utilized a deep neural network to predict the displacement from an input image patch to an anatomical landmark and employed a statistical shape model (SSM) to regularize the whole detection process. Arik et al. \cite{arik2017fully} trained a CNN on small patches to output probabilistic estimations of landmarks, and refined the positions of landmarks by a shape-based model. Xu et al. \cite{2017Supervised} leveraged an FCN to estimate action map (up, down, left, or right) and localized the landmark location from the estimated action map by a robust aggregation approach. Lee et al. \cite{lee2017cephalometric} trained 38 independent CNNs to regress the coordinates of the 19 cephalometric landmarks separately, and their method is very time-consuming for training and testing. Wu et al.~\cite{2017Automatic} extended CNN by a robust feature embedding layer to remove outlier features and a structured multi-output regression layer to regress landmark coordinates. 

However, direct regression of the coordinates from images involves a highly nonlinear mapping, which has been observed by research works for human pose estimation~\cite{toshev2014deeppose,pfister2015flowing}.
Therefore, many regression-based landmark detection methods predict heatmap images, which encode the pseudo-probability of a landmark being located at a certain pixel position. Payer et al.~\cite{payer2016regressing} used CNN to regress the heatmap of each landmark and used another network to combine the local features of landmarks with their spatial relations to all other landmarks to improve the prediction accuracy. O’Neil et al.~\cite{o2018attaining} trained an FCN with low-resolution images to learn spatial context and subsequently trained another FCN with higher resolution images and learned spatial information for further refinement.
Payer et al. \cite{2019Integrating} combined U-Net and their SpatialConfigurationNet by multiplying their output heatmap predictions for accurate and robust landmark detection.
Zhong et al. \cite{zhong2019attention} proposed two-stage U-Nets for landmark detection. A global U-Net takes an entire image as input and regress the heatmaps of landmarks in low-resolution. Guided by the coarse attention from the global stage, a local stage with patch-based U-Net regresses heatmaps in high-resolution. 
Chen et al. \cite{chen2019cephalometric} proposed an attentive feature pyramid fusion module to fuse features from different levels of a pre-trained network, then combined predicted heat maps and offset maps to perform pixel-wise regression-voting to improve detection accuracy. 
DACFL\cite{oh2020deep} forces the CNN to learn richer representations by perturbing the local appearance of training images based on prior anatomical distribution, and adopts the Anatomical Context loss to help learning the anatomical context based on spatial relationships between the landmarks.

Some works transfers landmark detection to other tasks such as objection detection and image segmentation. 
Qian et al. \cite{qian2019cephanet} detected landmarks by Faster R-CNN with a multi-task loss function and use a two-stage repair strategy to remove the abnormal candidate landmarks. 
Liu et al. \cite{liu2020misshapen} converted the detection of a landmark to the segmentation of the landmark's local neighborhood and solved it with a UNet-based approach which employs a non-local module with pyramid sampling to capture the global structural features.

To alleviate the problem of limited training data for anatomical landmark detection, our network utilizes a backbone network pre-trained on nature images whose feature extraction capacity is more powerful than the shallow UNet-based networks.
The work~\cite{chen2019cephalometric} also utilizes a pre-trained network to extract multi-scale features and enhances the fused feature with attention to improve the prediction accuracy. However, the main drawback of this work is that the size of its attention enhanced feature maps is linear to the number of landmarks which greatly increases the number of parameters, memory storage, and the computational cost.

\subsection{Multi-Scale Feature Fusion}
\par To achieve accurate and robust landmark detection, local features and context features should be combined. Multi-scale feature fusion aims to aggregate features at different resolutions. In the last a few years, some multi-scale feature aggregation networks have been proposed for object detection~\cite{lin2017feature,liu2018path}, image segmentation~\cite{long2015fully,ronneberger2015u,chen2017deeplab} and human pose estimation~\cite{newell2016stacked,xiao2018simple}. Among them, the encoder-decoder structure is widely-used where an encoder module contains a down-sampling convolution path to extract the semantic and context information from the input image and a decoder module has an up-sampling convolution path to recover spatial information of features. Skip connections from encoder layers to decoder layers with the same resolution are often used to preserve spatial information at each resolution. Some encoder-decoder networks such as U-Net~\cite{ronneberger2015u} and Hourglass~\cite{newell2016stacked} are shallow networks that limit their capacities. A solution to this is to stack multiple such networks as in~\cite{newell2016stacked}, but it remarkably increases the number of parameters and model size. Other networks such as FCN~\cite{long2015fully}, FPN~\cite{lin2017feature}, DeepLab~\cite{chen2017deeplab}, and the simple baseline network~\cite{xiao2018simple} use pre-trained network like VGG, ResNet, etc. as their encoders for feature extraction, and an up-sampling path as decoder to combine multi-scale features. 
PANet~\cite{liu2018path} adds an extra down-sampling feature aggregation path on top of FPN to enhance the entire feature hierarchy with accurate localization signals. Tan et al.\cite{tan2020efficientdet} proposed BiFPN which treats each bidirectional (up-sampling and down-sampling) path as a feature network layer and repeat it multiple times to enable more high-level feature fusion. In the work~\cite{chen2019cephalometric} for cephalometric landmark detection, $1\times1$ lateral connections and up-sampling are applied to features from different level of the backbone network to generate feature maps with the same resolution and number of channels. Then, these feature maps are concatenated together and passed through a dilated convolutional layer to fuse them.

\begin{figure*}[htbp]
\centering
\includegraphics[scale=0.56]{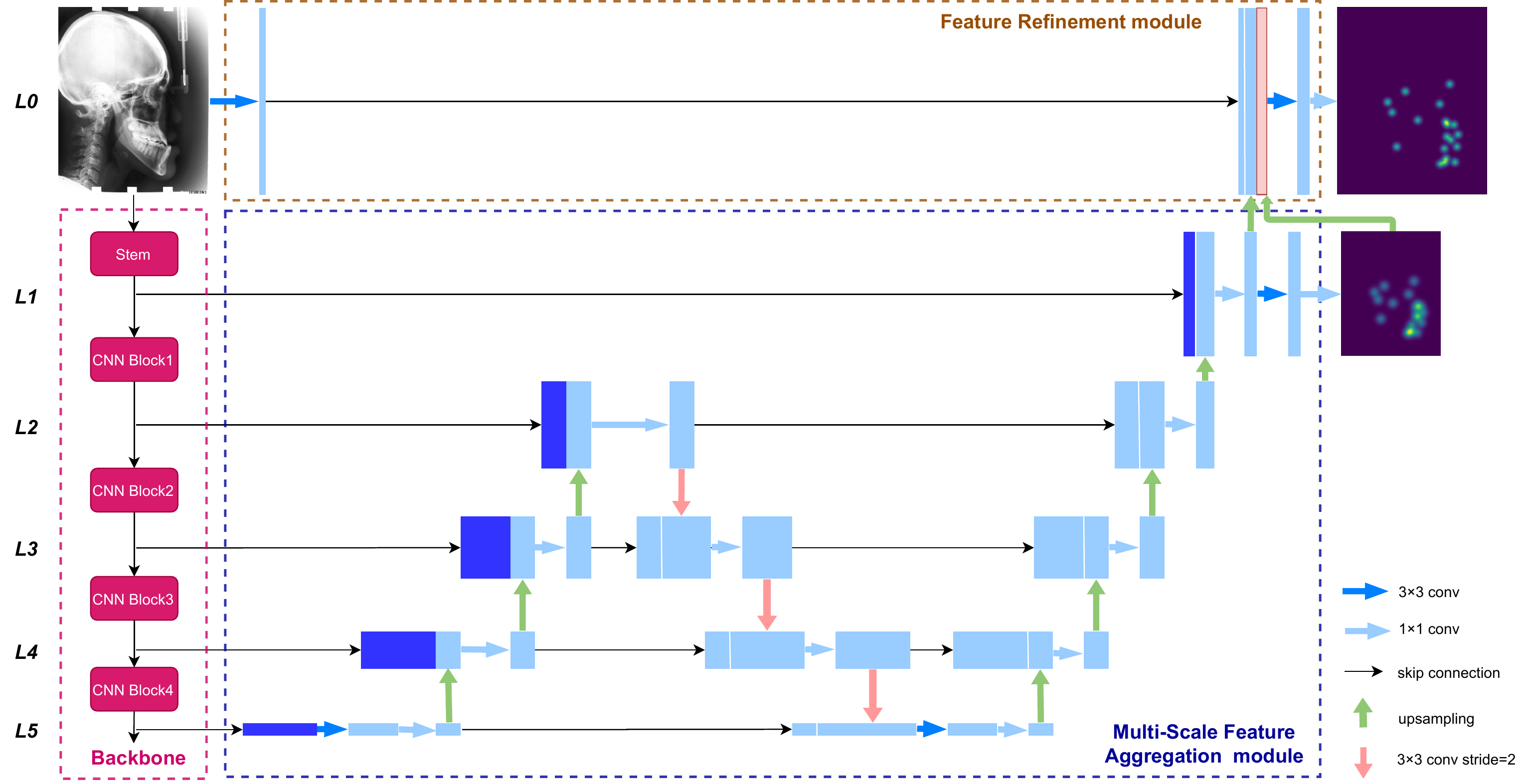}
\caption{The architecture of the feature aggregation and refinement network (FARNet). FARNet includes a backbone network (in the pink dashed box), a multi-scale feature aggregation (MSFA) module (in the blue dashed box) and a feature refinement (FR) module (in the brown dashed box). We also give the feature level labels $\left\{L0, L1, L2, L3, L4, L5\right\}$ at the left side of the figure, and all feature maps at the same horizontal level have the same spatial resolution. } 
\label{farnet} 
\end{figure*}

\par Inspired by FPN and its variations, we propose a multi-scale feature aggregation (MSFA) module. To make a good trade-off between the network capacity and efficiency, our MSFA module has one bidirectional (up-sampling and down-sampling) path followed by an up-sampling path. Compared to FPN, our MSFA module has one more down-sampling and up-sampling paths, compared to PAN and Bi-FPN, our network has one more up-sampling path to produce high-resolution feature maps, compared to the repeated Bi-FPN used in EfficientDet, our MSFA module is more efficient. On the other hand, these previous networks utilize the feature maps from convolution stage 2 to stage 5 of the backbone network, and our MSFA module uses the feature maps output from all stages and can generate finer heatmaps. Moreover, previous methods~\cite{lin2017feature, liu2018path, tan2020efficientdet} fuse features by addition, while our network combines features by concatenation which is more effective and flexible. 

\subsection{Loss Functions for Heatmap Regression}
Heatmap regression has become the mainstream approach to anatomical landmark detection, and Mean Square Error (MSE) or $L_2$ loss is used as the loss function. In face alignment, Adaptive Wing loss~\cite{wang2019adaptive} is proposed for robust heatmap regression. The Wing loss~\cite{feng2018wing} is a modified log loss for direct regression of landmark coordinates, which is able to pay more attention to the samples with small or medium range errors. The Adaptive Wing loss updates it to focus more on loss from foreground pixels than background pixels. In this paper, we proposed a simpler and effective loss function for heatmap regression. Our loss function is based on the $L_2$ loss and update it by multiply a factor to focus more on loss from foreground pixels than background pixels. Our experiments indicates the new loss function is more effective than the Adaptive Wing loss for anatomic landmark detection.

\section{Method}

\subsection{Feature Aggregation and Refinement Network}
{\color{Cerulean}Fig.} \ref{farnet} shows the overall architecture of our feature aggregation and refinement network, FARNet, which has a backbone network, a multi-scale feature aggregation (MSFA) module, and a feature refinement (FR) module. The backbone network adopts a popular ImageNet-pretrained network, such as VGG~\cite{simonyan2014very}, ResNet~\cite{he2016deep}, or DenseNet~\cite{huang2017densely}, etc. The backbone computes a feature hierarchy consisting of feature maps at multiple scales. Our MSFA module and FR module adopt up-sampling, down-sampling feature aggregation paths and lateral connections to fuse multi-scale features extracted by the backbone and provide high-resolution feature maps for landmark prediction. 
FPN uses the feature maps output by last residual block of convolution stage 2,3,4,and 5 of ResNet, and denotes them as $\left\{C2, C3, C4, C5\right\}$ which have strides of $\left\{4, 8, 16, 32\right\}$ pixels with respect to the input image.
The stem block in ResNet includes a $7\times7$ convolutional layer with stride 2 followed by a max-pooling layer with stride 2, and we denote the feature maps output from the convolutional layer of the stem block as $C1$ which have strides of 2 pixels with respect to the input image. For convenience, we denote the feature levels $\left\{L0, L1, L2, L3, L4, L5\right\}$ which  have strides of $\left\{0, 2, 4, 8, 16, 32\right\}$ pixels with respect to the input image.
To achieve high-resolution prediction, our MSFA module combines features from $C1$ to $C5$, and our FR module further combines features with the same resolution as the input image, which is denoted as $C0$. Note that, FPN and PANet only combine features from $C2$ to $C5$, and Bi-FPN takes features from $C3$ to $C7$ of EfficientNets. Compared to them, our network can use higher resolution features, which could be helpful to accurate landmark detection. More details about the MSFA module and FR module are given in the following.

\subsubsection{Multi-scale Feature Aggregation Module}
\par The backbone network extracts feature maps from the input image at several scales with a scaling step of 2 and works as the first down-sampling path. The feature maps output from the last layer of each convolution stage of the backbone network are input to our MSFA module. The MSFA module includes a bidirectional (up-sampling and down-sampling) path followed by an up-sampling path to combine the multi-scale features. In each feature fusion block, features with different resolutions are combined by concatenation in a higher-resolution-dominate manner, where the features with higher resolution are emphasized by keeping more channels for them than for those with lower resolution.

\begin{figure}[htbp]
\centering                                                          
\subfigure[]
{                    
    \begin{minipage}{4cm}
    \centering                                                          
    
    \includegraphics[scale=0.5]{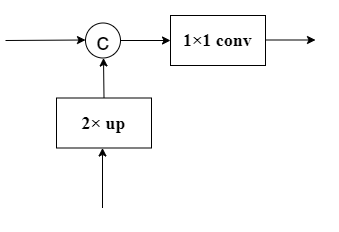}                
    \label{up-sampling}
    \end{minipage}
}    
\subfigure[]
{                    
    \begin{minipage}{4cm}
    \centering                                                          
    \includegraphics[scale=0.5]{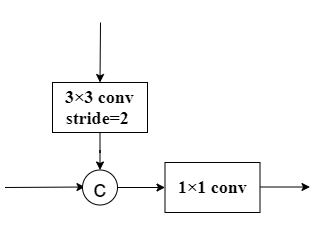}               
    \label{down-sampling}
    \end{minipage}
    
}\caption{Building block for feature aggregation. (a) a building block of up-sampling path combines feature maps from up-sampling path and lateral connection by concatenation and a $1\times 1$ convolution, (b) a building block of down-sampling path combines feature maps from down-sampling path and lateral connection by concatenation and a $1\times 1$ convolution.}                         
\label{fig:1}                                                        
\end{figure}

{\color{Cerulean}Fig.} \ref{up-sampling} shows the building block for the up-sampling path in MSFA module. In the block, the coarser-resolution feature maps are up-sampled by a factor of 2 and concatenated with the feature maps with the same resolution from the previous down-sampling path, and the concatenated feature maps go through a $1\times1$ convolution layer to reduce the number of channels to 256. The coarsest feature maps (at the feature level $L5$) from the previous down-sampling path are processed by a $3\times3$ and a $1\times1$ convolutional layers to produce the feature maps to the start of the up-sampling path. In our work, the first up-sampling path only reaches the feature layer $L2$, which is the feature level of the second stage of the backbone, and the second up-sampling path reaches the feature level $L1$, which is the feature level of the first stage of the backbone. Since the the resolution of feature maps gets larger while reaching the lower feature level, it's better to use small channel size to reduce the parameters. The second up-sampling path reduces the number of channels to 128 at $L2$ level, and reduces it to 64 at $L1$. At the end of the second up-sampling path, a $3\times3$ convolution and a $1\times1$ convolution are performed over the combined feature maps to regress heatmaps.

{\color{Cerulean}Fig.} \ref{down-sampling} shows the building block for the down-sampling path in MSFA module. In this block, the finer-resolution feature maps are down-sampled by a $3\times3$ convolutional layer with stride 2 and the number of channels is doubled to compensate for the loss of information caused by the decrease of resolution. The down-sampled feature maps are concatenated with the feature maps with the same resolution from the previous up-sampling path and the channel number is compressed back to the one before concatenation by a $1\times1$ convolutional layer. The output feature maps (at the feature level $L2$) from the first up-sampling path are used as the finest feature maps to the start of the down-sampling path. 

In previous methods~\cite{lin2017feature,liu2018path,tan2020efficientdet}, multi-scale features are merged by addition. In the up-sampling path, an up-sampling operation and a $1\times1$ convolution are used for matching both the resolution and the channel size before the addition. And in the down-sampling path, a down-sampling operation is used for matching both of them. After the addition, a $3\times 3$ convolution is usually used for feature processing. In our network, the multi-scale feature fusion is conducted by concatenation followed by a $1\times1$ convolutional layer. The concatenation operation allows different channel sizes for features with the different scales, which is more flexible and effective. The effectiveness of our multi-scale feature fusion is demonstrated by our experiments.

\subsubsection{Feature Refinement Module}                                        
The feature maps output from MSFA have the half resolution as the input image. To achieve more accurate prediction, our feature refinement (FR) module is introduced to generate feature maps having the same resolution as the input image. In the FR module, a $3\times3$ convolutional layer is performed over the input image, and the result feature maps (32 channels) are concatenated with the up-sampled feature maps (64 channels) and heatmaps from the MSFA module. The concatenated feature maps have the same resolution as the input image, to best of our knowledge, which is the highest resolution used among the related works. The heatmaps from MSFA module are used to guide the heatmap regression in higher resolution (FR module). Finally, we also perform a $3\times3$ convolutional layer and a $1\times1$ convolutional layer over the concatenated features to regress heatmaps. Through experimental verification, our feature refinement module can effectively improve the accuracy of prediction.

\subsection{Ground-Truth}
For heatmap regression, the ground-truth heatmaps are generated by applying an unnormalized Gaussian kernel to each landmark location. The ground-truth value $\textbf{H}_k(i,j)$ at $(i,j)$ in the heatmap for landmark $k$ is defined as following,
\begin{equation}\label{eq1}
     \textbf{H}_k(i,j)=\text{exp}\left(-\frac{(i-i_k)^2+(j-j_k)^2}{2\sigma^2}\right)
\end{equation}
where $(i_k,j_k)$ is the ground-truth position of landmark $k$ in the image, and $\sigma$ controls the spread of the peak. The heat map represents the pseudo-probability or confidence of a landmark being located at a certain pixel position. CNN is trained to fit the ground-truth heatmaps, and the coordinate regression is transferred to the heatmap regression. At test time, the coordinates of candidate landmarks are recovered by performing non-maximum suppression (NMS) over the predicted heatmaps.

Before encoding coordinates into heatmaps, the original image needs to be down-sampled to the input size of CNN. Accordingly, the ground-truth joint coordinates are usually downsampled and quantized to get the coordinates in heatmaps. And the heatmaps generated in this way are inaccurate due to the quantization error. To alleviate this problem, we follow \cite{zhang2020distribution} to use the non-quantized coordinates to generate more accurate heatmaps.

\subsection{Exponential Weighted Center Loss for Heatmap Regression}

\par In previous methods for heatmap regression, Mean Square Error (MSE) loss is commonly used like following,
\begin{equation}\label{eq2}
    \begin{aligned}
    Loss = \frac{1}{KWH}\sum\limits_{k=1}^K\sum\limits_{i=1}^W\sum\limits_{j=1}^HL_2(y_{i,j,k},\hat{y}_{i,j,k})
    \end{aligned}
\end{equation}
where $L_2(y,\hat{y})=(y-\hat{y})^2$ is the $L_2$ loss, $\hat{y}_{i,j,k}$ and $y_{i,j,k}$ are the pixel's intensities at position $(i,j)$ in the predicted heatmap and the ground truth heatmap of landmark $k$ respectively, and $W$ and $H$ are the width and height of heatmaps. The loss for an image is the average of the $L2$ loss over pixels in heatmaps of all landmarks, and all pixels have the same weight in the function.

The essence of heatmap regression-based landmark detection is to predict heatmap close to an unnormalized Gaussian distribution centered at each ground-truth landmark and use NMS to determine the landmark's coordinates. The regression accuracy at pixels near a landmark is more important for accurate localization of landmarks. On the contrary, the prediction accuracy at pixels far from a landmark is less important, since small errors on these pixels will not affect landmark localization. Therefore, the loss function should be able to adapt to the intensities of the pixels on the ground-truth heatmaps.

\par
\begin{figure}[htbp]
\centering
\includegraphics[scale=0.5]{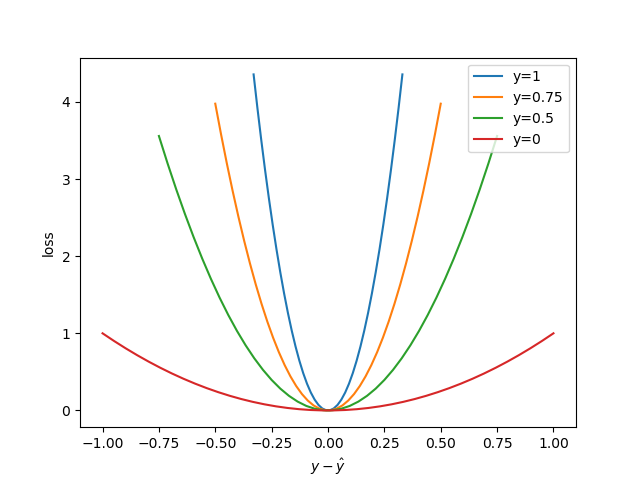}
\caption{The EWC loss function ($\alpha = 40$).} 
\label{EWC Loss} 
\end{figure}

 Based on the analysis above, we proposed a novel loss function named Exponential Weighted Center loss for heatmap regression, which is defined as follows:
\begin{equation}\label{eq6}
    EWC(y,\hat{y})=(y-\hat{y})^2\alpha^{y}
\end{equation}
where $\alpha$ is a hyper-parameter. From this equation, we can see that the error at every pixel in a heatmap is weighted by an exponential function of the ground-truth intensity $y$ at that pixel. The EWC loss function is also illustrated in {\color{Cerulean}Fig.} \ref{EWC Loss}. When $y$ is equal to 1, the weight reaches the maximum of $\alpha$ at the position of a landmark, and the loss is heavily enlarged. When the position moves away from a landmark, $y$ reduces and approaches 0, and the weight reduces exponentially to 1 with the intensity $y$. Therefore, more attention is paid to the errors at the pixels near a landmark and less attention is paid to the error at the pixels far away from it. In other words, the loss function focuses on the errors near a landmark and is less sensitive to the errors from the background area in an image. In our study, we set $\alpha$ to 40 to get good performance.

\subsection{Coarse-to-Fine Supervision}
In~\cite{szegedy2015going}, it is indicated that intermediate supervision plays an essential role in improving the performance of a deep neural network. For more accurate landmark prediction, we also introduce intermediate supervision to the MSFA module. Both the output feature maps and heatmaps of our MSFA module have 1/2 resolution as the input image, so we set its ground-truth heat maps to the same resolution. The predicted heatmaps of our FR module has the same resolution as the input image, and so does its ground-truth heatmaps. We set the kernel size $\sigma$ of the ground-truth heatmaps for the two modules to the same value, which means that the ground-truth for the MSFA module are coarser than those for the FR module and these two ground-truths lead to multi-scale coarse-to-fine supervision. Our experimental results demonstrate that the coarse-to-fine supervision is able to refine the localization.

\section{EXPERIMENTS AND RESULTS}
\subsection{Datasets}
In this paper, we evaluate our landmark detection network on three public benchmark data sets, a cephalometric X-rays dataset~\cite{wang2016benchmark}, a hand X-rays dataset~\cite{2019Integrating} and a Spinal Anterior-Posterior (AP) X-rays dataset~\cite{2017Automatic}. The example of these three datasets and their typical landmarks are shown in {\color{Cerulean}Fig.} \ref{example}. 

\subsubsection{Cephalometric X-rays}
The cephalometric X-rays dataset is provided by the ISBI 2015 Grand Challenge in Automatic Detection and Analysis for Diagnosis in Cephalometric X-ray Images~\cite{wang2016benchmark}. It consists of 400 lateral cephalometric X-ray images from 400 different subjects, with 19 annotated landmarks labeled by two experienced doctors. Each image has a resolution of $1935\times2400$ and each pixel is about 0.1mm. The dataset is split into a training data set of 150 images, a Test1 dataset of 150 images, and a Test2 dataset of 100 images. We use the training data set for training, the Test1 dataset for validation, and the Test2 dataset for testing. We adopt the evaluation metrics used in the ISBI 2015 Challenge~\cite{wang2016benchmark}, which include the mean radial error (MRE, in mm, the smaller the better), defined as the error between manually and automatically marked landmarks, and the successful detection rate (SDR, the bigger the better) in radius (2.0mm, 2.5mm, 3.0mm, 4.0mm). We take the average of two doctors' annotations as ground truth.

\subsubsection{Hand X-rays}
\par The hand X-rays dataset contains 895 X-ray images of left hands with an average size of $1563\times2169$ pixels from a publicly available Digital Hand Atlas\footnote{Digital Hand Atlas Database System, \url{www.ipilab.org/BAAweb}}. In~\cite{2019Integrating}, the annotations of 37 characteristic landmarks on fingertips and bone joints are provided. Following~\cite{2019Integrating}, we normalize the image resolution according to wrist widths, and adopt the three-fold cross-validation setup which splitting images into approximately 600 training and 300 testing images per fold. The evaluation metrics include the mean radial error (MRE, in mm) and the successful detection rate (SDR) in radius (2mm, 4mm, 10mm).

\subsubsection{Spinal Anterior-Posterior X-rays}
The spinal AP X-rays dataset contains of 481 spinal anterior-posterior x-ray images provided by clinicians~\cite{2017Automatic}. 17 vertebrae composed of the thoracic and lumbar spine are selected for spinal shape characterization. Each vertebra is located by four landmarks at four corners thus resulting in 68 landmarks per spinal image. Following~\cite{2017Automatic}, the dataset is split into 431 for training/validation and 50 for testing. Since the authors of~\cite{2017Automatic} have not shared their data split, we split the data randomly. The evaluation metrics include the Mean Squared Error (MSE) and Pearson Correlation Coefficient ($\rho$) between the predicted landmarks and annotated ground truth.

\subsection{Implementation Details}
Our network is implemented by PyTorch 1.0.1 and Python 3.6. For the cephalometric X-rays dataset, the input image is resized to $800\times640$ and no data augmentation is conducted. For the hand X-ray dataset, the input image is resized to $512\times512$ and data augmentation is employed following~\cite{2019Integrating}. For the spinal AP X-ray dataset, the input image is resized to $1024\times512$ and data augmentation is performed following~\cite{2017Automatic}. Through experimental comparison, we set the kernel size $\sigma$ which is a parameter used to generate the ground-truth heatmaps to 10 and the hyper-parameter $\alpha$ to 40. The network is optimized by Adadelta optimizer and the learning rate is 0.0001. The backbone parameters are optimized along with the entire network. We train our network for 300 epochs on a GTX 2080TI GPU with a mini-batch size of 1. 

\subsection{Comparison of Backbone Networks}
We first conduct experiments on the Test1 data of the cephalometric X-ray dataset to compare several popular backbone networks, including VGG, ResNet, and DenseNet. ResNet and DenseNet have a similar structure as shown in {\color{Cerulean}Fig.} \ref{farnet}. VGG has five convolutional blocks corresponding to $\left\{ L0, L1, L2, L3, L4\right\}$, therefore our FR module directly combines the first block's output feature maps with the up-sampled feature maps and heatmaps from the MSFA module. As shown in {\color{Cerulean}Table} \ref{backbone}, DenseNet-121 has achieved the best performance, hence we adopt it in all the following experiments. On the contrary, VGGNets need 900 epochs to converge and obtain the worst performances.

\begin{table}
    \footnotesize
    \centering
    \caption{Comparison of different backbone networks on the Test1 data of the cephalometric X-ray dataset.}
    \begin{tabular}{c|c|c|c|c|c}
        \hline
         & MRE & 2mm & 2.5mm & 3mm & 4mm  \\
         \hline
         VGG-16 & 1.44 & 84.03 & 90.70 & 93.81 & 97.29\\
         VGG-19 & 1.37 & 82.31 & 89.08 & 92.98 & 96.87\\
         ResNet-101 & 1.19 & 86.49 & 92.28 & 95.40 & 98.07\\
         ResNet-152 & 1.29 & 86.76 & 92.42 & 95.33 & 98.03\\
         DenseNet-169 & 1.15 & 87.64 & 92.13 & 95.49 & 98.38 \\
         \textbf{DenseNet-121} & \textbf{1.12} & \textbf{88.03} & \textbf{92.73} & \textbf{95.96} & \textbf{98.48}  \\
         \hline
         
    \end{tabular}
    \label{backbone}
\end{table}

\subsection{Ablation Study}
To better understand the merit of each component in our framework, we conduct ablation studies to evaluate them on the Test1 data of cephalometric X-ray dataset. The components evaluated include our MSFA module, FR module, coarse-to-fine supervision, and the proposed Exponential Weighted Center (EWC) loss function. To validate our multi-scale feature fusion, we implement another version of the MSFA module, MSFA(+), which follows FPN~\cite{lin2017feature} and PAN~\cite{liu2018path} to merge features by addition and uses the same channel size setting. 
To evaluate the gain of coarse-to-fine supervision, we develop a na\"ive version (suffixed by *) of the FR module with no supervision on the MSFA and thus no heatmap from the MSFA introduced to the FR module. To validate our loss function, we also evaluate our network with the Adaptive Wing loss (AW). Furthermore, we compare our network with the popular U-Net~\cite{ronneberger2015u} and FPN~\cite{lin2017feature} to validate our backbone and MSFA module. 

The results in {\color{Cerulean}Table} \ref{ablation} show that original U-Net achieves the worst results among the comparing methods, and our network with only the backbone and the MSFA module (MSFA) can outperform it in all metrics by a large margin (MRE reduced by 0.21, SDR in 2.0mm improved by 1.72). This is mainly because U-Net is very shallow, and the pre-trained backbone used in our network can extract more powerful features and our MSFA module enable more high-level feature fusion. For a fair comparison, we also adopt DenseNet-121 as the backbone for FPN. After upsampling the finest feature maps from FPN to the resolution of the input image, a $3\times3$ and a $1\times1$ convolution layers are performed on them to regression heatmaps. From the experimental results we can see,  FPN is better than U-Net which is mainly due to the backbone used. MSFA(+) outperforms FPN which indicate one more down-sampling and up-sampling path can make a better feature fusion. And our MSFA module can further improve over MSFA(+), which is due to its more effective feature fusion strategy.

The na\"ive version of the FR module (FR*) reduces MRE by 0.01 and improves SDR in 2.0mm by 0.74. When applying coarse-to-fine supervision to the FR module and MSFA module and introducing the up-sampled heatmaps from MSFA to the FR module, the FR module further reduces MRE by 0.01 and improves SDR in 2.0mm by 0.52. This validates the use of coarse-to-fine supervision and the heatmap-guide strategy. Finally, when employing the Exponential Weighted Center loss in the supervsion, our FARNet reduces MRE by 0.03, and improves SDR in 2.0mm by 0.6, and achieves the best performance. In our study, the Adaptive Wing loss can not improve over the $L_2$ loss, and our EWC loss function can beat it at all metrics. These results indicate that the proposed components can consistently improve the accuracy of landmark localization.

\par
\begin{figure*}[htbp]
\centering  
\includegraphics[scale=0.5]{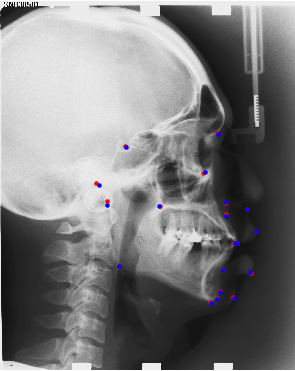}  
\hspace{1ex}									
\includegraphics[scale=0.5]{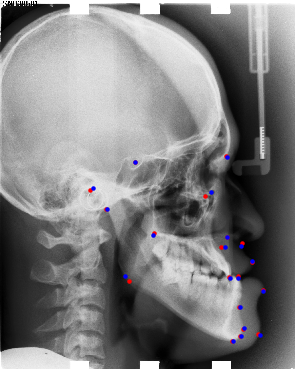} 
\hspace{1ex}									
\includegraphics[scale=0.5]{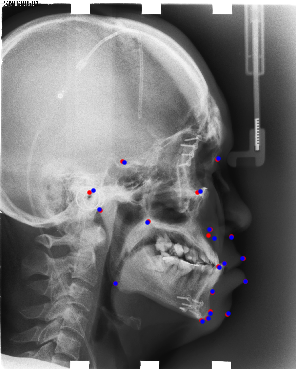} 
\hspace{1ex}									
\includegraphics[scale=0.5]{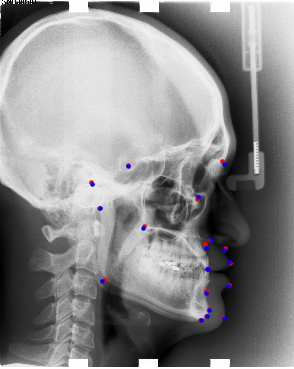} \\
\includegraphics[width=110pt,height=160pt]{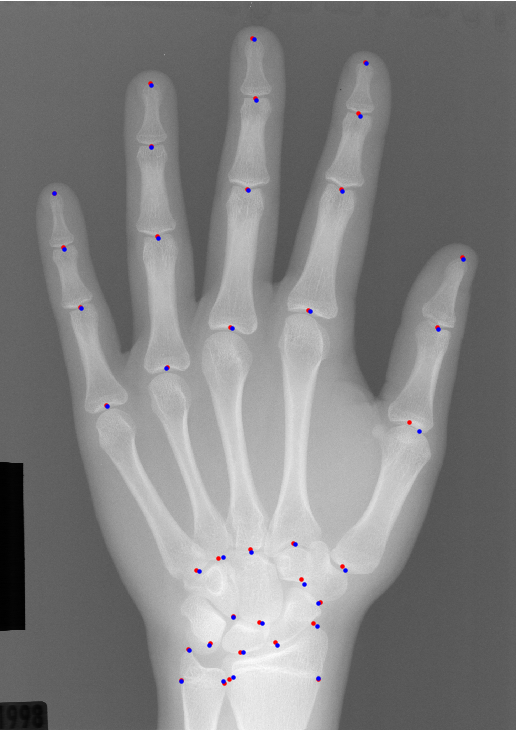}
\hspace{1ex}
\includegraphics[width=110pt,height=160pt]{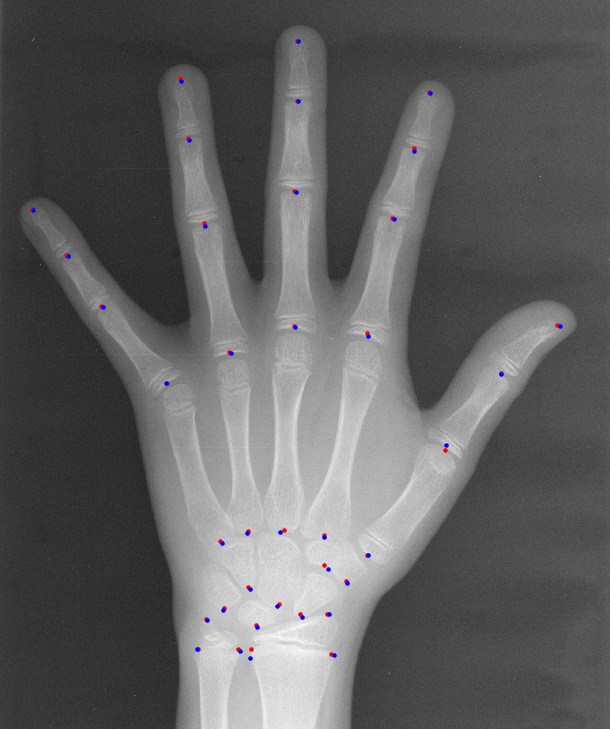}
\hspace{1ex}
\includegraphics[width=110pt,height=160pt]{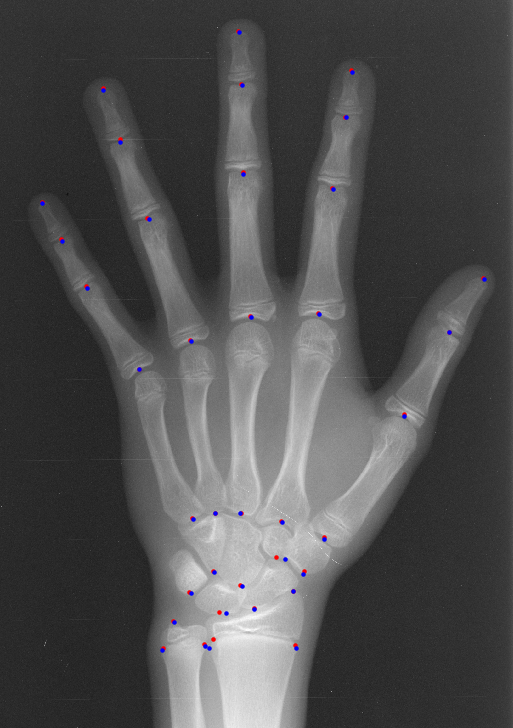}
\hspace{1ex}
\includegraphics[width=110pt,height=160pt]{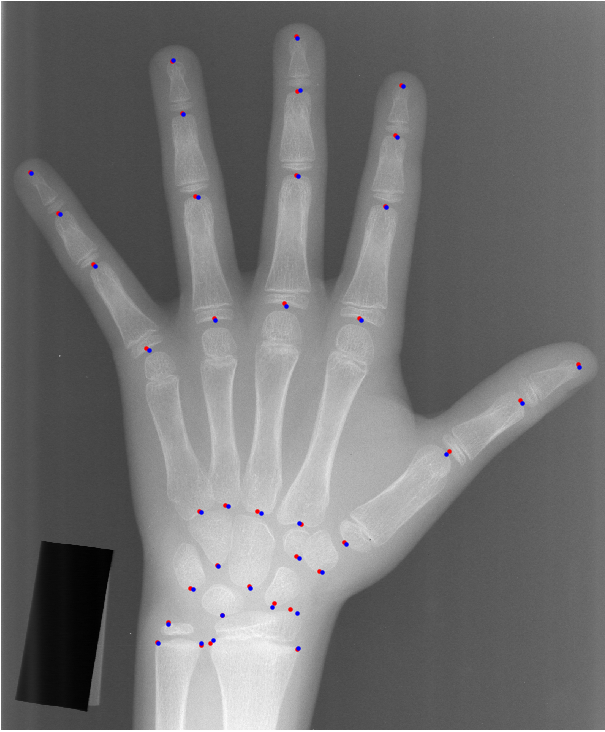} \\
\includegraphics[width=85pt,height=210pt]{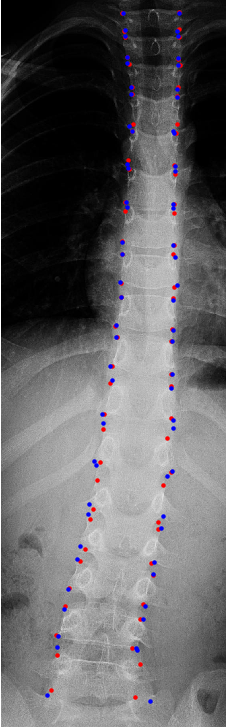}
\hspace{1ex}
\includegraphics[width=85pt,height=210pt]{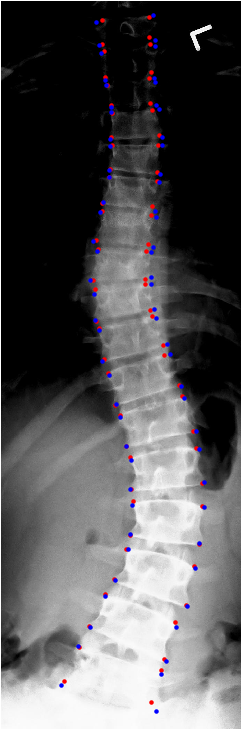}
\hspace{1ex}
\includegraphics[width=85pt,height=210pt]{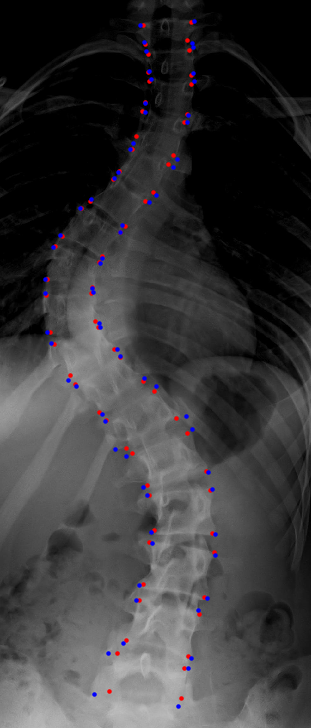}
\hspace{1ex}
\includegraphics[width=85pt,height=210pt]{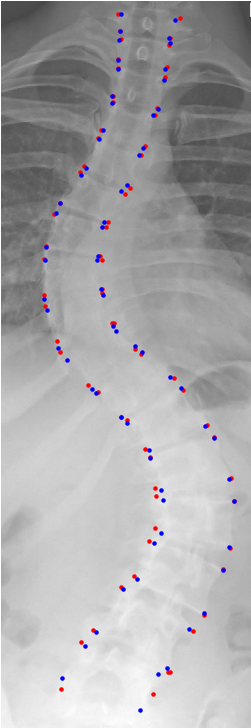} 
\hspace{1ex}
\includegraphics[width=85pt,height=210pt]{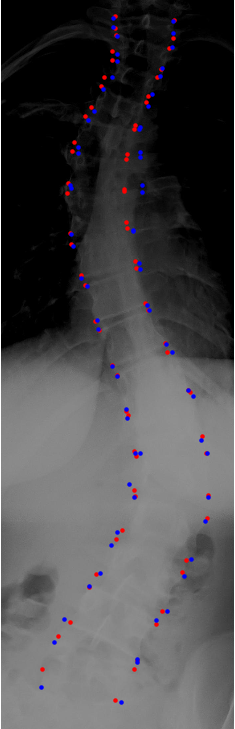} 
\caption{Illustration of landmark detection results by our proposed method on three public datasets. The first row is the task of cephalometric landmark detetcion(19 landmarks), the second row is the task of hand radiographs landmark detection(37 landmarks) and the last row is the task of spinal anterior-posterior x-ray landmark detection(68 landmarks). The red points denote our detected landmarks via our framework, while blue points represent the ground-truth landmarks.}
\label{results}
\end{figure*}

 {\color{Cerulean}Fig.} \ref{results} shows some representative results by our network on the three data sets. The red points denote the landmarks detected by our network, and the blue points represent the ground-truth landmarks.

\begin{table}
    \footnotesize
    \centering
    \caption{Ablation study: the MSFA module, na\"ive FR module, coarse-to-fine supervision, and the proposed Exponential Weighted Center (EWC) loss function.}
    \begin{tabular}{c|c|c|c|c|c}
        \hline
         & MRE & 2mm & 2.5mm & 3mm & 4mm  \\
         \hline
         UNet & 1.38 & 84.45 & 90.45 & 93.57 & 97.33\\
         FPN & 1.19 & 85.47 & 92.17 & 95.54 & 98.24\\
         MSFA(+)  & 1.18 & 85.73 & 92.31 & 95.83 & 98.36\\
         MSFA & 1.17 & 86.17 & 92.42 & 95.64 & 98.38\\
         MSFA+FR* & 1.16 & 86.91 & 92.63 & 95.68 & 98.45 \\
         MSFA+FR & 1.15 & 87.43 & \textbf{93.01} & 95.85 & 98.45 \\
         MSFA+FR+AW & 1.15 & 87.08 & 92.31 & 95.64 & 98.38 \\
         \textbf{MSFA+FR+EWC} & \textbf{1.12} & \textbf{88.03} & 92.73 & \textbf{95.96} & \textbf{98.48} \\
         \hline
        
    \end{tabular}
     \label{ablation}
\end{table}

\subsection{State-of-the-art Comparison}
\subsubsection{Cephalometric X-rays dataset}
We first compare our method with prior state-of-the-art methods on the cephalometric X-rays dataset. All the experimental results on Test1 data and Test2 data are shown in {\color{Cerulean}Table} \ref{compare cepha}.
\begin{table*}[ht]
    \centering
    \normalsize
    \caption{Comparison of our FARNet with prior state-of-the-art methods on the cephalometric X-ray dataset with 19 annotated landmarks.}
    \label{tab1}
    \begin{tabular}{|c|c|c|c|c|c|c|c|c|c|c|c|}
        \hline
        \multirow{2}*{Methods}&  \multirow{2}*{Input size} &
        \multicolumn{5}{|c|}{Test1 data} &  \multicolumn{5}{|c|}{Test2 data}\\
        \cline{3-12}
         & & MRE & 2mm & 2.5mm & 3mm & 4mm & MRE & 2mm & 2.5mm & 3mm & 4mm\\
        \hline
        Ibragimov et al.\cite{ibragimov2015computerized} &- &1.84	&71.70	&77.40	&81.90	&88.00	&-	&62.74	&70.47	&76.53	&85.11\\
        \hline
        Lindner et al.\cite{lindner2015fully}	&- &1.67	&74.95	&80.28	&84.56	&89.68	&1.92	&66.11	&72.00	&77.63	&87.42\\
        \hline
        Arik et al.\cite{arik2017fully} &$800\times640$ &- &75.37 &80.91 &84.32 &88.25 &- &67.68 &74.16 &79.11 &84.63 \\
        \hline
        Qian et al.\cite{qian2019cephanet} &- &- &82.50	&86.20	&89.30	&92.60	&-	&72.40	&76.15	&79.65	&85.90\\
        \hline
        Oh et al.\cite{oh2020deep}	&$800\times640$ &1.18	&86.20	&91.20	&94.40	&97.70	&1.44	&75.89	&83.36	&89.26	&9\textbf{5.73}\\
        \hline
        Chen et al.\cite{chen2019cephalometric}	&$800\times640$ &1.17	&86.67	&92.67	&95.54	&\textbf{98.53}	&1.48	&75.05	&82.84	&88.53	&95.05\\
        \hline
        Zhong et al.\cite{zhong2019attention} &$968\times968$ &1.12 &86.91 &91.82 &94.88 &97.90 &1.42 &76.00 &82.90 &88.74 &94.32\\
        \hline
        \textbf{FARNet(Our)}	&\textbf{$800\times640$} & \textbf{1.12} & \textbf{88.03} & \textbf{92.73} & \textbf{95.96} & 98.48 	&  \textbf{1.42} &\textbf{77.00}	&\textbf{84.42}	&\textbf{89.47}	& 95.21 \\
        \hline
    \end{tabular}
    \label{compare cepha}
\end{table*}

\begin{table*}[ht]
    \centering
    \normalsize
    \caption{Landmark localization results from a three-fold cross validation on the Hand X-rays dataset with 37 annotated landmarks and compare with other methods. }
\begin{tabular}{|c|c|c|c|c|c|}
        \hline
        Methods & Input size & MRE $\pm$ Std(mm) & 2mm(\%) & 4mm(\%) & 10 mm(\%) \\
        \hline
        Urschler et al\cite{urschler2018integrating} & $1250\times1250$ & 0.80 $\pm$ 0.93 & 92.19 & 98.46 & 99.95 \\
        \hline
        Štern et al\cite{vstern2016local} & $1250\times1250$ & 0.80 $\pm$ 0.91 & 92.20 & 98.45 & 99.95 \\
        \hline
        Ebner et al\cite{ebner2014towards} & $1250\times1250$ & 0.97 $\pm$ 2.45 & 91.60 & 97.84 & 99.31 \\
        \hline
        Lindner et al\cite{lindner2014robust} & $1250\times1250$ & 0.85 $\pm$ 1.01 & 93.68 & 98.95 &99.94 \\
        \hline 
        Payer et al\cite{2019Integrating} & $512\times512$ & 0.66 $\pm$ 0.74 & 94.99 & 99.27 & 99.99 \\
        \hline
        \textbf{FARNet(Our)} & \textbf{$512\times512$} & \textbf{0.62 $\pm$ 0.55} &\textbf{97.24} &\textbf{99.8}	&\textbf{100}\\
        \hline
    \end{tabular}
    \label{compare hand}
\end{table*}

\par Ibragimov et al. \cite{ibragimov2015computerized} and Lindner et al. \cite{lindner2015fully} combined the random forests regression-voting and the statistical shape analysis techniques, and have achieved the best performances in the IEEE ISBI 2014~\cite{wang2015evaluation} and 2015 Challenges~\cite{wang2016benchmark} respectively. Ibragimov's method obtains the MRE of 1.84 mm on Test1 data and the SDRs of 71.70\% and 62.74\% on Test1 and Test2 data respectively using 2mm precision range, which is the acceptable precision range in clinical practice. 
In the following description, we only mention the SDRs in this range. Lindner's method makes an improvement and achieves the MRE of 1.67 mm on Test1 data and the SDRs of 74.95\% and 66.11\% on Test1 and Test2 data respectively. Arik et al. \cite{arik2017fully} combined a CNN with a shape-based model for landmark detection, and their method achieves the SDRs of 75.37\% and 67.68\% on Test1 and Test2 data respectively. Qian et al. \cite{qian2019cephanet} utilized Faster R-CNN to detect landmarks and a two-stage repair strategy to remove the abnormal candicate landmarks. Their method makes a remarkable improvement over previous methods, and achieves the SDRs of 82.50\% and 72.40\% on Test1 and Test2 data respectively. DACFL\cite{oh2020deep} learns richer representations and achieves the SDRs of 86.20\% and 75.89\% on Test1 and Test2 data respectively. Chen et al. \cite{chen2019cephalometric} combined multi-scale features from a pre-trained backbone network and further employed a self-attention mechanism for landmark detection. Their method makes further improvement and achieves the SDRs of 86.67\% and 75.05\% on Test1 and Test2 data respectively. But, their method utilized a self-attention mechanism to construct weighted feature maps for different landmarks separately, which greatly increases the number of parameters and the consumption of memory storage. 
Zhong et al. \cite{zhong2019attention} first utilized a global U-Net to regress coarse heatmaps from a downsized image and utilize the heatmaps to guide a patch-based U-Net to regress heatmaps in high resolution. Their method achieves the SDRs of 86.91\% and 76.00\% on Test1 and Test2 data respectively. However, their patch-based prediction strategy and Expansive Exploration refine strategy greatly increase the training and testing time.

Finally, our FARNet achieves the best results on all the evaluation metrics. It obtains 1.12 and 1.00 points improvements of SDRs on Test1 and Test2 data respectively over the second-best method \cite{zhong2019attention}. And our network is more efficient in training and testing.    


\subsubsection{Hand X-rays}
To evaluate our deep network on the hand X-rays dataset, we follow the standards of Payer et al \cite{2019Integrating} and also use three-fold cross-validation. We compare our method with their method which achieve the best results recently and also with other prior state-of-art methods \cite{urschler2018integrating,vstern2016local,ebner2014towards,lindner2014robust}. The results are shown in {\color{Cerulean}Table} \ref{compare hand}. The prior state-of-art methods are mainly random forest-based approaches. Among them, Lindner et al\cite{lindner2014robust} obtained the best SDR of 93.68\% in 2mm precision range, and Štern et al\cite{vstern2016local} achieved the best MRE of 0.80mm. Payer et al\cite{2019Integrating} combined U-Net with a learned global configuration for landmark localization and greatly improve the performance. Their method obtains the SDR of 94.99\% and the MRE of 0.66mm. Our FARNet remarkably improves the performance and achieves the SDR of 97.24\% and the MRE of 0.62mm. 

\subsubsection{Spinal Anterior-Posterior X-rays}
In this experiment, we evaluate our FARNet on the public spinal anterior-posterior x-ray dataset\cite{2017Automatic} and compare our method with BoostNet\cite{2017Automatic} and other baseline methods on this dataset. We conduct 5-fold cross-validation on the Trainset and evaluate them on the Test data. The mean square error (MSE) and Pearson Correlation Coefficient ($\rho$) are used as the evaluation metrics. The unit for MSE is a fraction of the original image (e.g. 0.010 MSE represents an average of 10-pixel error in a $100\times100$ image). The experimental results are shown in {\color{Cerulean}Table} \ref{compare spine}.  From it, we can see that our method outperforms previous methods by large margins, which proves its effectiveness and generality.

\begin{table}[htbp]
    \centering
    \normalsize
    \caption{Landmark localization results on the Spinal Anterior-Posterior X-ray dataset with 68 annotated landmarks and compare with other methods. The units of MSE is the fraction of orinal image (0.010 MSE represents average of 10pixel error in a $100\times100$ image)}
\begin{tabular}{|c|c|c|}
        \hline
        Methods & MSE(fraction of image) & $\rho$ \\
        \hline
        SVR\cite{sun2017direct} & 0.006 & 0.93 \\
        \hline
        RFR\cite{criminisi2010regression} & 0.0052 & 0.94 \\
        \hline
        BoostNet\cite{2017Automatic} & 0.0046 &0.94 \\
        \hline
        \textbf{FARNet(Our)} & \textbf{0.0017} & \textbf{0.98}\\
        \hline
    \end{tabular}
    \label{compare spine} 
\end{table}

\section{Conclusion}
In this paper, we propose a novel end-to-end deep network for anatomical landmark detection. Our network includes a backbone network, a feature aggregation module and a feature refinement module. The backbone network pre-trained on natural images is used to extract a feature hierarchy. The feature aggregation module is used to fuse multi-scale features extracted by the backbone network, and the feature refine module is adopted to generate high-resolution feature maps. Coarse-to-fine supervisions are applied to the two modules to facilitate the end-to-end training. We further propose a novel loss function for more accurate heatmap regression, which concentrates on the losses from the pixels near landmarks and suppresses the ones from far away. Our network has achieved state-of-art performances on three publicly available anatomical landmark detection datasets, which demonstrates the effectiveness and generality of our network. And the end-to-end nature of our network make it more efficient than the previous patch-based approaches.

\bibliographystyle{unsrt}
\bibliography{cas-refs}

\end{document}